\def\year{2021}\relax
\pdfoutput=1
\documentclass[letterpaper]{article} 
\usepackage{aaai21}  
\usepackage{times}  
\usepackage{helvet} 
\usepackage{courier}  
\usepackage[hyphens]{url}  
\usepackage{graphicx} 
\usepackage{graphicx}  
\usepackage{natbib}  
\usepackage{caption} 
\usepackage{cite}
\usepackage{subfigure} 
\usepackage{multirow}
\usepackage{booktabs}
\usepackage{amssymb}
\usepackage{amsmath}

\title{Towards Robust Visual Information Extraction in Real World: \\New Dataset and Novel Solution}
\author{

    Jiapeng Wang\textsuperscript{\rm 1},
    Chongyu Liu\textsuperscript{\rm 1},
    Lianwen Jin\textsuperscript{\rm 1,3,}\thanks{Corresponding author},
    Guozhi Tang\textsuperscript{\rm 1},
    Jiaxin Zhang\textsuperscript{\rm 1},
    Shuaitao Zhang\textsuperscript{\rm 1},
    Qianying Wang\textsuperscript{\rm 2},
    Yaqiang Wu\textsuperscript{\rm 2,4},
    Mingxiang Cai\textsuperscript{\rm 2}
    \\
}
\affiliations{

    \textsuperscript{\rm 1}South China University of Technology;
    \quad \textsuperscript{\rm 2}Lenovo Research; \\
    \textsuperscript{\rm 3}SCUT-Zhuhai Institute of Modern Industrial Innovation; 
    \quad \textsuperscript{\rm 4}Xi'an Jiaotong University\\
    \{eejpwang,
    eechongyu.liu\}@mail.scut.edu.cn,
    eelwjin@scut.edu.cn,
    \{eetanggz,
    eejxzhang,
    eestzhang\}@mail.scut.edu.cn,
    \{wangqya,
    wuyqe,
    caimx\}@lenovo.com\\
}

\begin{document}

\maketitle

\begin{abstract}
Visual information extraction (VIE) has attracted considerable attention recently owing  to its various advanced applications such as document understanding, automatic marking and intelligent education. Most existing works decoupled this problem into several independent sub-tasks of text spotting (text detection and recognition) and information extraction, which completely ignored the high correlation among  them during optimization. In this paper, we propose a robust visual information extraction system (VIES) towards real-world scenarios, which is an unified end-to-end trainable framework for simultaneous text detection,  recognition and information extraction by taking a single document image as input and outputting the structured information.  Specifically, the information extraction branch collects abundant  visual and semantic representations from text spotting for multimodal feature fusion and conversely,  provides higher-level semantic clues to contribute to the optimization of text spotting. Moreover, regarding  the shortage of public benchmarks, we construct a fully-annotated dataset called EPHOIE (https://github.com/HCIILAB/EPHOIE),
which is the first Chinese benchmark for  both text spotting and visual information extraction. EPHOIE  consists of 1,494 images of examination paper head with complex layouts and background, including a total of 15,771 Chinese handwritten or printed text instances.
Compared with the state-of-the-art methods, our VIES shows significant   superior performance on the EPHOIE dataset and achieves a  9.01\%  F-score gain on the widely used SROIE dataset under the end-to-end scenario.
\end{abstract}

\section{Introduction}
Recently, visual information extraction (VIE) has attracted considerable research  interest owing to its various advanced applications, such as document understanding \cite{wong1982document}, automatic marking \cite{tremblay2003semi}, and intelligent education \cite{kahraman2010development}.

Most existing works for VIE mainly comprise  two independent stages, namely text spotting  (text detection and recognition) and information extraction. The former aims to locate and recognize the texts, while the latter extracts specific entities based on previous results.
Recent studies \cite{GCN,PICK,Layoutlm} revealed that in addition to semantic features, the visual and spatial characteristics of documents also provided abundant  clues. Although achieved encouraging results, these approaches still suffered from the following limitations: (1) Although their text spotting models had learned effective representations for  detection and recognition, their information extraction modules discarded and then retrieved them again from the OCR results. This resulted in redundant computation, and the discarded features might be more effective than the newly learned ones. (2) The training processes of independent parts were irrelevant, leading to the lack of clues obtained by the information extraction module, while the text spotting module cannot be optimized adaptively according to the fundamental objective. Continuous stages were usually combined  to accomplish a common task, while they did not collaborate with each other.

To address the limitations mentioned above, in this paper, we propose a robust visual information extraction system towards real-world scenarios called VIES, which is an unified end-to-end trainable framework for simultaneous text detection, recognition and information extraction. VIES  introduces  vision coordination mechanism (VCM)  and  semantics coordination mechanism (SCM)  to gather rich visual and semantic features from text detection and recognition branches respectively for subsequent information extraction branch and conversely, provides higher-level semantic clues to contribute to the optimization of text spotting. 
Concurrently,  a novel  adaptive feature fusion module      (AFFM)   is designed to integrate the features from different sources (vision, semantics and location) and levels (segment-level and token-level) in information extraction branch to generate more effective representations. 

With the vigorous development of learning-based algorithms, a comprehensive benchmark conducted for a specific task is a prerequisite to motivate more advanced works. In VIE, SROIE \cite{sroie} is the most widely used one,
which concentrates both on the optical character recognition (OCR) and VIE tasks for scanned receipts in printed English. However, it's difficult to satisfy the demand of real-world applications for documents with complex layouts and handwritten texts.

\begin{figure}[t!]
\centering
\subfigure[Complex Layout]{
\begin{minipage}[]
{0.45\linewidth}
\centering
\includegraphics[width=4cm,height=5cm]{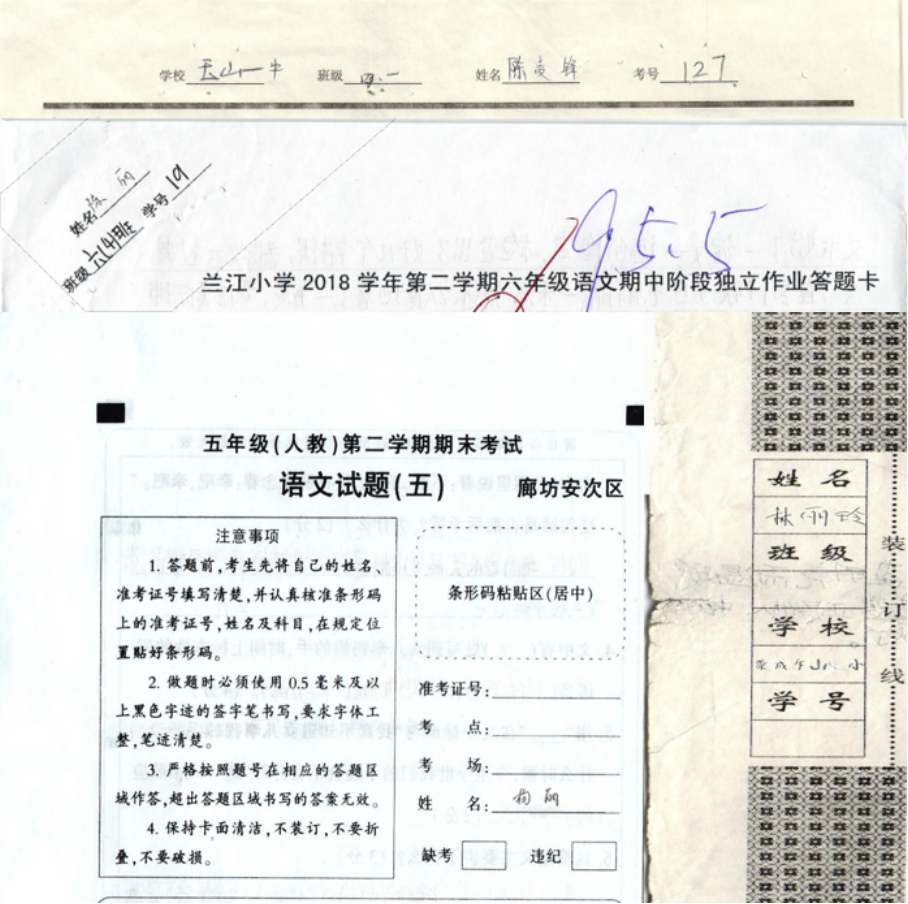}\\
\end{minipage}
}%
\subfigure[Noisy Background]{
\begin{minipage}[]
{0.45\linewidth}
\centering
\includegraphics[width=4cm,height=5cm]{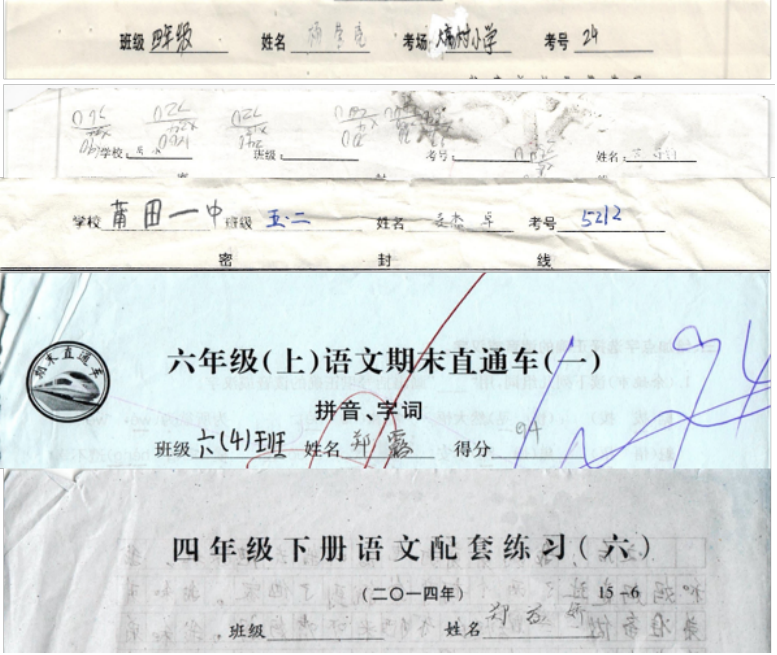}\\
\end{minipage}
}%
\caption{Some typical and challenging instances in EPHOIE. (a) Complex layout. (b) Noisy background.}\label{fig:data}
\end{figure}

To address this issue and promote the development of the field of VIE, we furthermore establish a challenging dataset called Examination Paper Head Dataset for OCR and Information Extraction (EPHOIE), which contains 1,494 images with 15,771 annotated text instances. The images are collected and scanned from real examination papers of various schools in China, and we crop the paper head regions which contains all key information. 
The texts are composed of handwritten and printed Chinese characters in horizontal and arbitrary quadrilateral shape. Complex layouts and noisy background also enhance the generalization of EPHOIE dataset. 
Typical examples are shown in Figure \ref{fig:data}. 

Extensive experiments on both the EPHOIE and widely used benchmarks demonstrate that our proposed VIES outperforms the state-of-the-art methods substantially.
Our main contributions can be summarized as follows:
\begin{itemize}
\item We propose a robust visual information extraction system towards real-world scenarios called VIES, which is an unified end-to-end trainable framework for simultaneous text detection, recognition and information extraction.
\item We introduce VCM and SCM to enable  the independent modules to benefit from joint optimization. AFFM is also designed to integrate features from different sources and levels to boost the entire  framework.
\item We propose a fully-annotated dataset called EPHOIE, which is the first Chinese benchmark for applications of both text spotting and visual information extraction.
\item Our method achieves state-of-the-art performance on both the EPHOIE  and widely  used benchmarks, which fully demonstrated the effectiveness of the proposed VIES.
\end{itemize}

\section{Ralated Work}
\subsubsection{Datasets for Visual Information Extraction}
For VIE, SROIE \cite{sroie} is the most widely used public dataset that has brought great impetus to this fields. 
It concentrates on scanned receipts in printed English, and contains complete OCR annotations and key-value pair information labels for each image.  \cite{Eaten} proposed a Chinese benchmark with fixed layouts including train tickets, passports and business cards. However, the overwhelming majority of images were totally synthetic and only annotated with key-value pair labels without any OCR annotations. 
In this regard, for the development of both OCR and VIE tasks in Chinese documents and handwritten information, a comprehensive dataset with complex background, changeable layouts and diverse text styles towards real-world scenarios is in great demand.

\begin{table*}[]
\caption{\label{tab:number} Comparison between EPHOIE and SROIE. `H'  or `Q' denotes horizontal or arbitrary quadrilateral text. }
\centering
 \resizebox{170mm}{8mm}{
\begin{tabular}{@{}cccccccc@{}}
\toprule
\textbf{Dataset} & \textbf{Year} & \textbf{Scenario}           & \textbf{Language} & \textbf{Image Number} & \textbf{Text Shape} & \textbf{Script}         & \textbf{Entities} \\ \midrule
\textbf{SROIE}    & 2019 & Scanned receipts   & English  & 973          & H          & Printed             & 4                  \\
\textbf{EPHOIE}   & 2020 & Scanned paper head & Chinese  & 1494         & H, Q       & Printed/Handwritten & 10                 \\ \bottomrule
\end{tabular}
}

\end{table*}

\subsubsection{Methods for Visual Information Extraction}
In recent years, VIE methods have achieved encouraging improvement. Early works mainly used rule-based \cite{c1,c2} or template matching \cite{c4} methods, which might led to the poor generalization. With the development of deep learning, more researchers  converted the results obtained by text spotting into plain texts, and then extracted feature embeddings for a subsequent sequence labeling model such as BiLSTM-CRF \cite{blstmcrf} to obtain the final entities. However, the lack of visual and location information often led to poor performance.

The fact that visual and spatial features of documents also play a vital role in information extraction has been recognized by recent works. Typical methods such as Post-OCR parsing  \cite{Post-OCR} took bounding box coordinates into consideration. LayoutLM  \cite{Layoutlm}  modeled the layout structure and visual clues of documents based on the pre-training process of a BERT-like model. GraphIE \cite{Graphie}, PICK \cite{PICK} and  \cite{GCN} tried to use Graph Neural Networks (GNNs) to extract global graph embeddings for further improvement. CharGrid \cite{Chargrid} used CNNs to integrate semantic clues contained in input matrices and the layout information simultaneously. However, these existing traditional methods only focused on the performance related to the information extraction stage, but ignored the preconditioned OCR module.

At present, more related works of VIE were gradually developing towards end-to-end manner.  \cite{Eaten} generated feature maps directly from input image and used several entity-aware decoders to decode all the entities. However, it could only process documents with fixed layout and its efficiency could be significantly reduced as the number of entities increases.  \cite{c18} localized, recognized and classified each text segment in image, which was difficult to handle the situation where a text segment was composed of characters with different categories.  \cite{TRIE} proposed an end-to-end trainable framework to solve VIE task. However, it focused more on the performance of entity extraction and can only be applied to the scenarios where the OCR task was relatively simple.

\section{Examination Paper Head Dataset for OCR and Information Extraction}
In this section, we introduce the new Examination Paper Head Dataset for OCR and Information Extraction (EPHOIE) benchmark and its characteristics.

\begin{figure}[t]
\centering
\subfigure[]{
\begin{minipage}[]{0.95\linewidth}
\centering
\includegraphics[width=8cm,height=3cm]{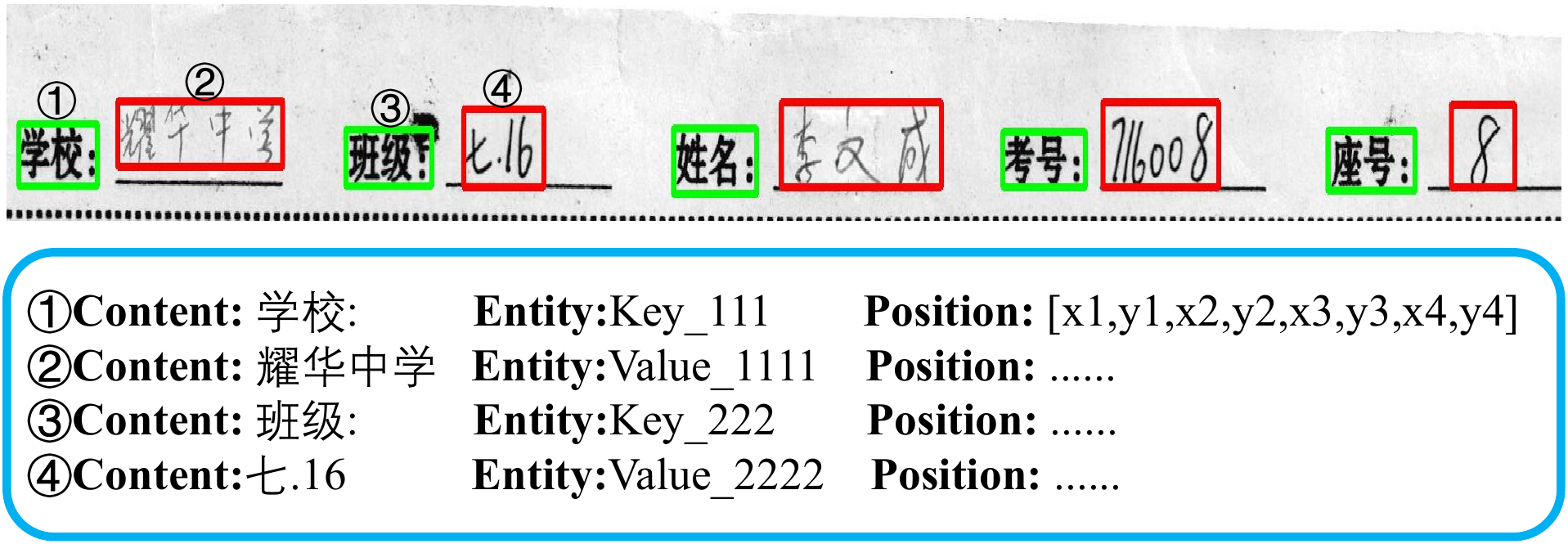}\\
\end{minipage}
}%
\quad
\subfigure[]{
\begin{minipage}[]{0.95\linewidth}
\centering
\includegraphics[width=8cm,height=3.5cm]{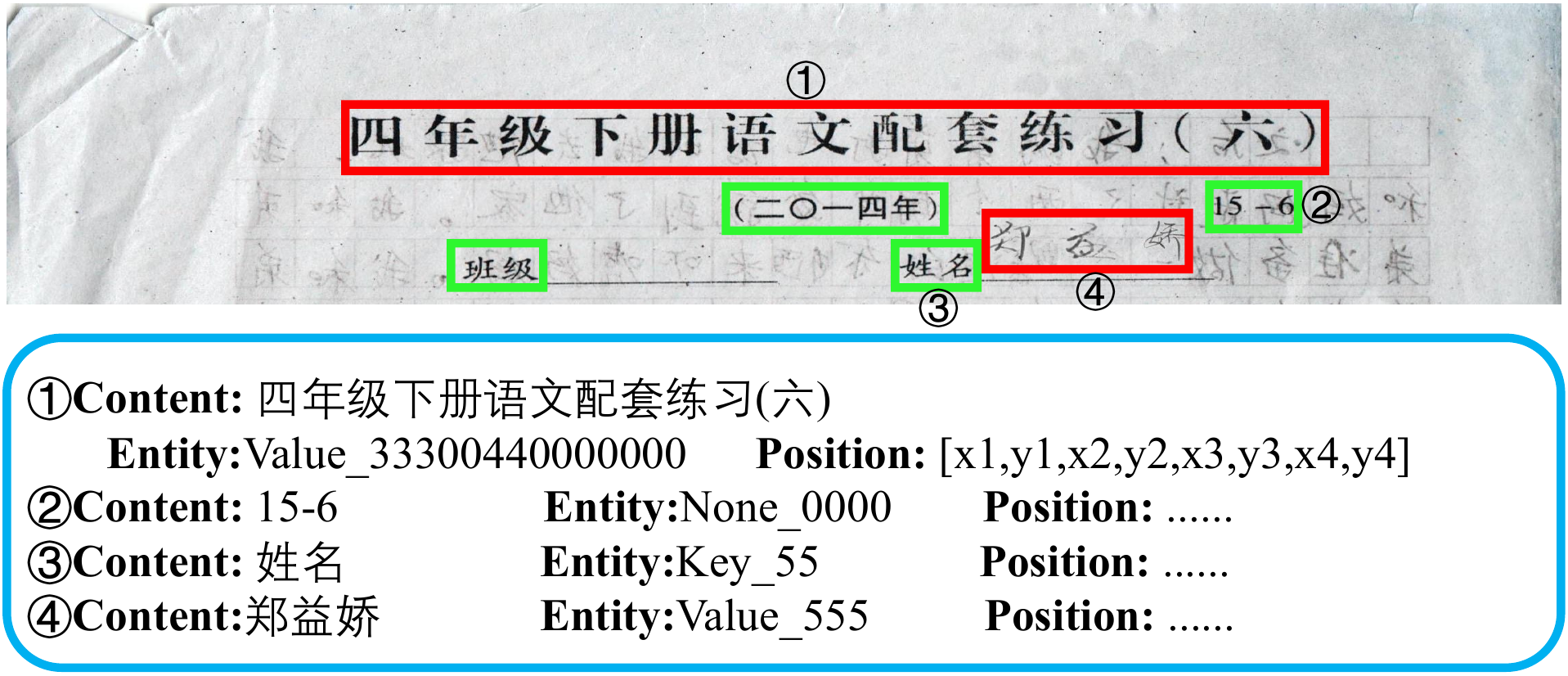}\\
\end{minipage}
}%
\caption{Examples of annotations in EPHOIE. In \textit{Entity} field, `Key' or `Value' indicates it's key or value of an entity respectively, whereas `None' indicates neither of them. The different numbers in \textit{Entity} denotes different categories. }\label{fig:annotation}
\end{figure}

\subsubsection{Dataset Description}
To the best of our knowledge, the EPHOIE benchmark  is the first public dataset for both OCR and VIE tasks in Chinese and aims to motivate more advanced works in the fields of both document intelligence and VIE. It contains 1,494 images with 15,771 annotated text instances, including handwritten and printed characters. It is divided into a training set with 1,183 images and a testing set with 311 images respectively. All the images in EPHOIE are collected and scanned from real examination papers of various schools with the diversity of text types and layout distribution. The statistic of our dataset and the comparison with the most widely used public benchmark SROIE are shown in Table \ref{tab:number}. For EPHOIE, we only crop the paper head regions that contain all key information. 

\begin{figure*}[t]
\centering
\includegraphics[width=1.0\textwidth]{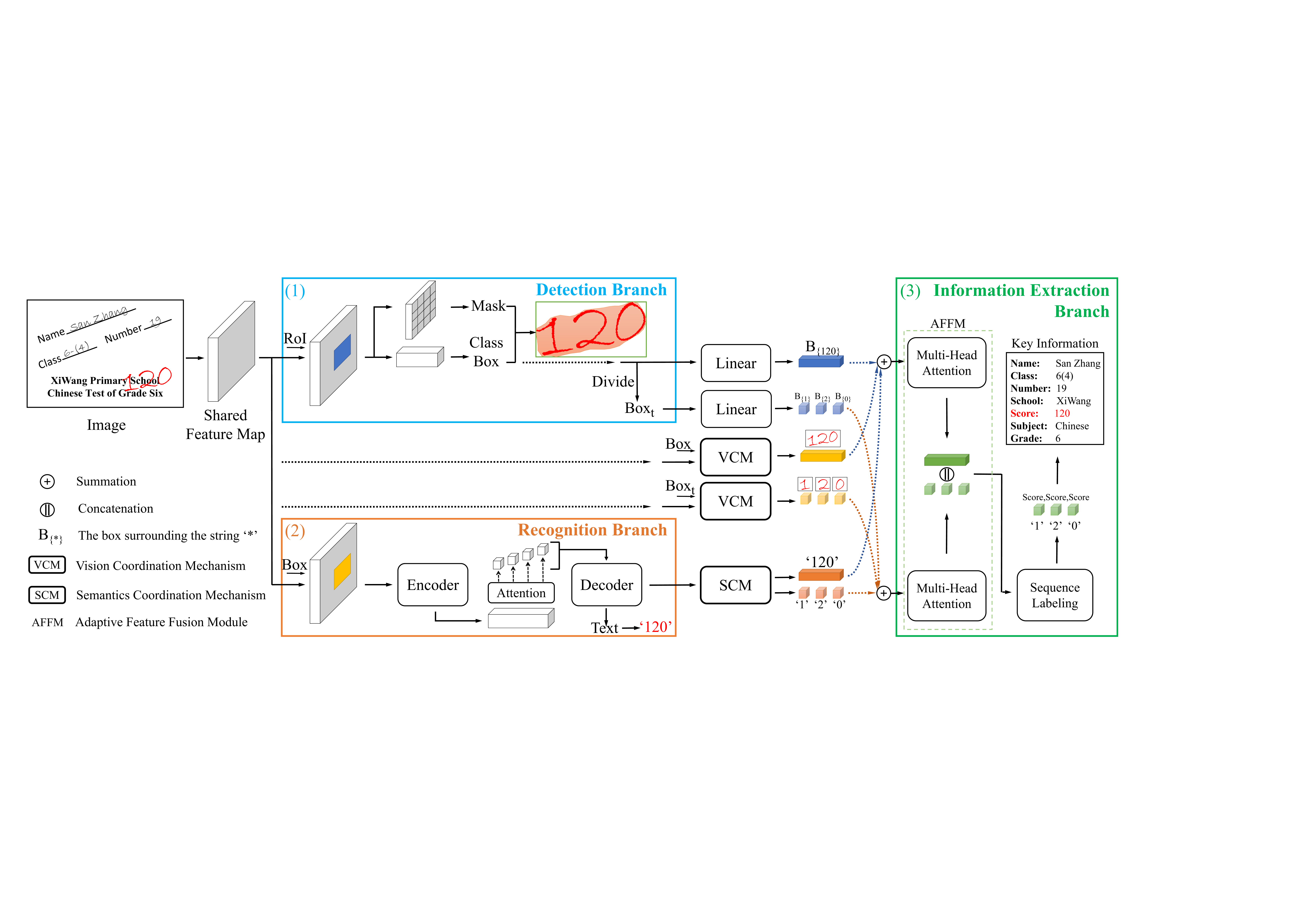} 
\caption{The overall framework of VIES. It  consists  of  a  shared  backbone  network  and  three  specific branches: (1) text detection, (2) text recognition and (3) information extraction. $\mathop {Box_t}$ denotes boxes of single tokens which divided from the box of entire text segment $\mathop {Box}$. }
\label{pipeline}
\end{figure*}

\begin{figure}[t]
\centering
\subfigure[]{
\begin{minipage}[]{0.95\linewidth}
\centering
\includegraphics[width=8cm,height=2cm]{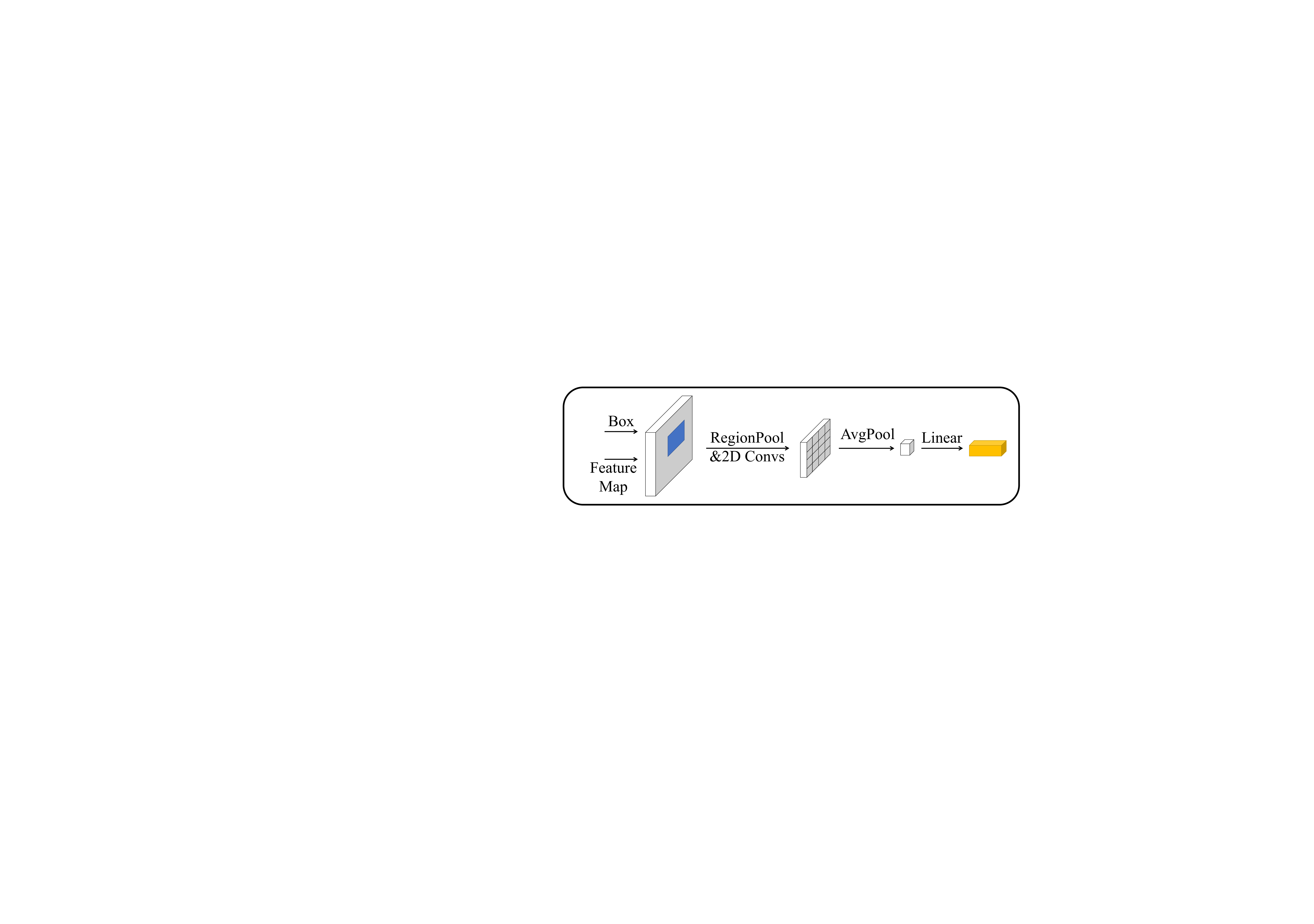}\\
\end{minipage}
}
\quad
\subfigure[]{
\begin{minipage}[]{0.95\linewidth}
\centering
\includegraphics[width=8cm,height=2cm]{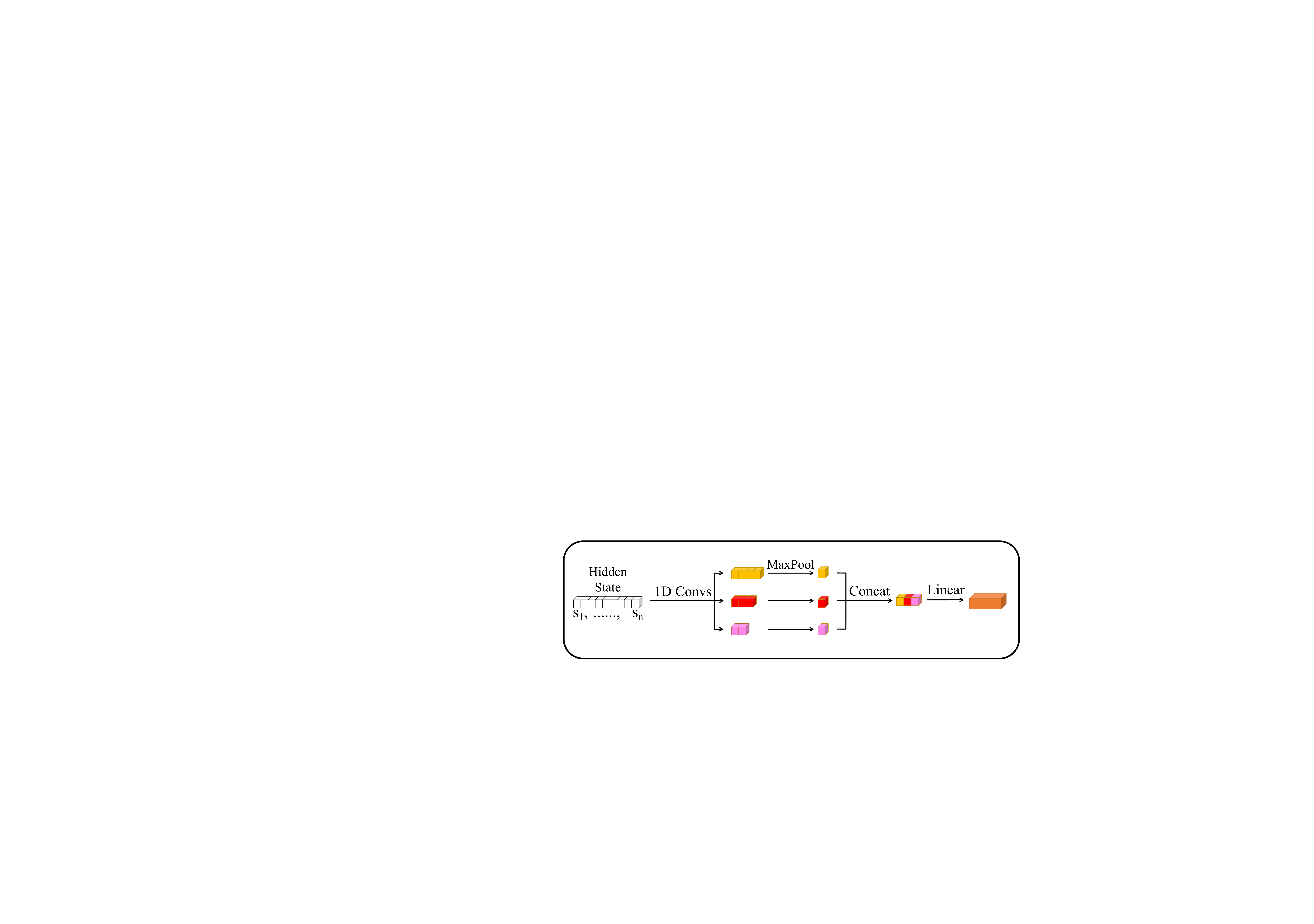}\\
\end{minipage}
}
\caption{The detailed structures of Vision and Semantics Coordination Mechanisms. (a) Vision Coordination Mechanism (VCM). (b) Semantics Coordination Mechanism (SCM). }
\label{fig:cm}
\end{figure}

\subsubsection{Annotation Details}

The detailed annotation forms of EPHOIE are presented  in Figure \ref{fig:annotation}. 
As there exist both horizontal and arbitrary quadrilateral texts, four vertices were required to surround them. In addition to annotating bounding boxes for text detection, text content is also required for both text recognition and information extraction. We annotate all the texts appearing on the image, while additionally label the entity key-value pair for all key information. 
The number string in \textit{Entity} denotes the category of each token, since there may exists multiple  entities in a single segment.

\section{Methodology}
The overall framework of our VIES is illustrated in  Figure \ref{pipeline}. It consists of a shared backbone and three specific branches of text detection, recognition and information extraction. Given an input image, the text spotting branches are responsible for not only localizing and recognizing all the texts in it, but also providing abundant  visual and semantic features through vision and semantics coordination mechanisms for subsequent networks. The adaptive feature fusion module in information extraction branch first gathers these abundant  representations with additional spatial features extracted from detected boxes to adaptively generate fused features in decoupled levels (segment-level and token-level). In this part, the multi-head self-attention mechanism is introduced to allow each individual to freely attend to all the others. Then, the features in decoupled levels are re-coupled and finally the specific entities are distinguished from recognized strings using sequence labeling module. 


\subsection{Text Detection with \\ Vision Coordination Mechanism (VCM)}

Accurate text detection is the prerequisite for  text recognition and information extraction. An intuitive idea to boost the detection branch is to make the IE branch provide feedback guidance in an end-to-end manner during training. 

Given an input image, our VIES first uses the shared backbone to extract high-level feature representations $\mathop X$. Then, the detection branch takes $\mathop X$ as input and outputs boxes $\mathop B$, confidence scores $\mathop C$ and even binary masks $\mathop M$ for arbitrary quadrilateral texts:
\begin{equation}
\mathop B, \mathop C, \mathop M  = \mathop {TextDetection}\nolimits_{} (\mathop X )
\end{equation}
Here, we introduce an innovative vision coordination mechanism (VCM), which can effectively transfer rich visual features $\mathop F\nolimits_{vis}$ from the detection branch to the IE branch and conversely provide additional supervised information to contribute to the optimization of the detection branch. It can be shown in Figure \ref{fig:cm}(a) and defined as follows:
\begin{small}
\begin{equation}
 \mathop F\nolimits_{vis}  = \mathop {Linear} ( \mathop {AvgPool} ( Conv2D ( \mathop {RegionPool} (\mathop X,\mathop B) )))
\end{equation}
\end{small}
Here, $\mathop {RegionPool}$ denotes region feature pooling methods such as RoIPooling \cite{fast} and RoIAlign \cite{mask}. $\mathop {AvgPool}$ reduces both height and width dimension to unit size. $\mathop {Linear}$ is learned projection to transform $\mathop F\nolimits_{vis}$   into $d$ channels.

For visually rich documents, the key visual clues such as shapes, fonts and colors have been integrated in $\mathop F\nolimits_{vis}$. The gradients of IE branch can also help the detection branch learn more general representations that are beneficial to the entire  framework.

\subsection{Text Recognition with \\ Semantics Coordination Mechanism (SCM)}
Text recognition greatly limits the upper bound of the performance of the entire   system. If the recognized strings are less accurate, it will always be useless no matter how powerful the IE branch is. Based on this consideration, whether semantic
supervision of IE stage can boost the recognition branch is particularly critical.

In our VIES, given the shared features $X$, high-level representations in specific text regions are collected and fed into an encoder to extract the input feature sequence $H = (\mathop h\nolimits_1 ,\mathop h\nolimits_2 , \cdots ,\mathop h\nolimits_N )$, where $N$ is the feature length. Then, an attention-based decoder \cite{attention} is adopted to recurrently generate the hidden states $S = (\mathop s\nolimits_1 ,\mathop s\nolimits_2 , \cdots ,\mathop s\nolimits_M ) $ by referring to the history of recognized characters and $H$, where $M$ indicates the maximum decoding step. Finally, the output text sequence $O = (\mathop o\nolimits_1 ,\mathop o\nolimits_2 , \cdots ,\mathop o\nolimits_M ) $ is computed using $S$.

Here, we introduce our semantics coordination mechanism (SCM) to establish the bidirectional semantics flow between our recognition branch and IE branch.
The hidden states $S$ in our recognition branch contain high-level semantic representations of each single token in every decoding step. 
Therefore, we regard it as token-level semantic features $\mathop F_{sem,t}$ and send it to the IE branch:
\begin{align}
 F_{sem,t} = ( s_1 , s_2 , \cdots , s_M )& =  S,\\
 where
 \quad
 F_{sem,t_i}  &= s_{i} \nonumber
\end{align}
Here, $\mathop F\nolimits_{sem,t_i}  \mathop { \in \mathbb{R}}\nolimits^d $ denodes the $d$-dimensional vector corresponding to the $i$-th token in the segment. 

Note that, the segment-level semantic features $\mathop F\nolimits_{sem,s}$ also greatly affect  the category characteristics. Further, $\mathop F\nolimits_{sem,t}$ captures local clues and $\mathop F\nolimits_{sem,s}$ contains global information, indicating that they are complementary to each other. 
Inspired by the previous works \cite{charnet,textcnn} which adopted CNNs to integrate holistic expression for each sentence from the words' or characters' embeddings, 
our VIES generates the summarization of each segment $\mathop F\nolimits_{sem,s}$ from $\mathop F\nolimits_{sem,t}$ as follows:
\begin{align}
\mathop F\nolimits_{sem,{t_{1:n}}}  & = \mathop F\nolimits_{sem,t{_1}}  \oplus  \cdots  \oplus \mathop F\nolimits_{sem,t{_n}}, \\
c_i & = Conv1D_i(F_{sem,t{_{1:n}}}), \\
c_i & = MaxPool1D(c_i),  \\ 
i & = 1, \ldots ,\mathop n\nolimits_c  \nonumber\\
F_{sem,s} & = Linear(c_{1}  \oplus  \cdots  \oplus c_{n_c}) 
\end{align}
Here, $\oplus$ is the concatenation operator, $\mathop n$ is the length of the current segment and $\mathop n\nolimits_c$ is the number of 1D convolution kernels. Note that all 1D operations are carried out over the \textit{length} dimension.

The overall structure of SCM is illustrated in Figure \ref{fig:cm}(b). 
In this way, the extracted competent semantic representations can be passed forward directly, and the higher-level semantic constraints of IE branch can guide the training process of recognition branch.


\subsection{Information Extraction with\\Adaptive Feature Fusion Module (AFFM)}
Information extraction  requires the most comprehensive and expressive representations to distinguish specific entities from recognized strings. Besides the visual and semantic features provided by text spotting branches above, our IE  branch further extracts spatial features from text boxes and  decouples token-level representations so that the relatively accurate  clues can be obtained regardless of whether the token is attributed to the wrong string or whether the string is over- or under-cut, which often occur in text detection owing to the complex background and the variety of shapes and styles. To encode location information, we generate spatial features $\mathop F\nolimits_{spt}$  from relative bounding box coordinates as:
\begin{equation}
\mathop F\nolimits_{spt}  = Linear([\frac{{\mathop x\nolimits_{min} }}{{\mathop W\nolimits_{img} }},\frac{{\mathop y\nolimits_{min} }}{{\mathop H\nolimits_{img} }},\frac{{\mathop x\nolimits_{max } }}{{\mathop W\nolimits_{img} }},\frac{{\mathop y\nolimits_{max } }}{{\mathop H\nolimits_{img} }}])
\end{equation}
where $\mathop W\nolimits_{img}$ and $\mathop H\nolimits_{img}$ are image width and height respectively. $\mathop {Linear}$ is used to transform $\mathop F\nolimits_{spt}$   into $d$ channels which is the same as that in the visual and semantic features above. We intuitively divide the box of the entire segment evenly along its longest side into several single tokens' boxes $\mathop B_{t}$ according to the length of recognized strings. 
Then the token-level visual features $\mathop F\nolimits_{vis,t}$ and spatial features $\mathop F\nolimits_{spt,t}$ can be generated  according to  $\mathop B_{t}$.

After acquiring the features of multi levels from multi sources as representations in a learned common embedding space, our adaptive feature fusion module (AFFM)  introduces two multi-layer multi-head self-attention modules combined with linear transforms to first enrich all projected vectors at different fine-granularities respectively. 
The multimodal features are summarized and followed by the layer normalization to generate comprehensive representations of each individual. Then, it serves as the $K$, $Q$ and $V$ in the scaled dot-product attention, which can expressed as:
\begin{gather}
Attention(Q,K,V) = softmax (\frac{{\mathop QK^T }}{{\sqrt {\mathop d } }})V,\\
where \quad Q , K ,V =  {LayerNorm(F}_{vis}  +  F_{sem}  +  F_{spt}) \nonumber
\end{gather}
\begin{gather}
 \mathop F\nolimits_{*}  = Concat(\mathop {head}\nolimits_1 ,\mathop {head}\nolimits_2 , \cdots ,\mathop {head}\nolimits_h )\mathop W\nolimits^O,\\
where \quad \mathop {head}\nolimits_i  =  Attention(\mathop {QW}\nolimits_i^Q ,\mathop {KW}\nolimits_i^K ,\mathop {VW}\nolimits_i^V )\nonumber
\end{gather}
where $\mathop W\nolimits_i^Q $, $\mathop W\nolimits_i^K $, $\mathop W\nolimits_i^V $ and $\mathop W\nolimits^O $ are projection parameters and $\mathop F\nolimits_{*}$ denotes the fused features in segment-level or token-level. Then, we re-couple them to combine global and local information:
\begin{align}
 \overline{F}_{j,t_{i}} & =  F_{j,t_{i}} \oplus  F_{j},\\ 
where \quad i & =1,...,n_{j},  \quad j=1,...,n_{s} \nonumber
\end{align}
where $n_s$ is the number of text segments and $n_j$ is the length of $j$-th segment. $\overline{F}_{j,t_i} $ constitutes input feature sequence for subsequent sequence labeling module.

\subsubsection{Sequence Labeling}
After feature recoupling, we feed the input feature sequence into standard BiLSTM-CRF \cite{blstmcrf} for entity extraction. Intuitively, segment embedding provides extra global representations. The concatenated features are fed into a BiLSTM network to be encoded, and the output is further passed to a fully connected network and then a CRF layer to learn the semantics constraints in an entity sequence.

\subsection{Optimization Strategy}
In the training phase, our proposed framework can be trained in an end-to-end manner with the weighted sum of the losses generated from three branches of text detection, recognition and information extraction:
\begin{equation}
L = \mathop L\nolimits_{E}  + \mathop {\mathop \lambda \nolimits_D L}\nolimits_D  + \mathop \lambda \nolimits_R \mathop L\nolimits_R 
\end{equation}
where $\lambda_D$ and $\lambda_R$ are hyper-parameters that control the tradeoff between losses.  $L_D$ and $L_R$ are losses of text detection and recognition branches respectively, and $L_{E}$ is the loss of information extraction branch.

$\mathop L\nolimits_{D}$ consists of losses for text classification, box regression and mask identification respectively, as defined in  \cite{mask}. $\mathop L\nolimits_{R}$ adopts CrossEntropyLoss between output text sequence $O$ and ground truth text sequence. CRFLoss is also adopted as $\mathop L\nolimits_{E}$ for information extraction.



\section{Experiments}
\subsection{Implement Details}
We adopt Mask R-CNN \cite{mask} as our text detection branch with ResNet-50 \cite{resnet} followed by FPN \cite{fpn} as its backbone. 
We use LSTM \cite{lstm} in attention mechanism for text recognition. In SCM, the sizes of three 1D convolutions  are 2, 3 and 4. In AFFM, we set the number of heads and sub-layers is 4 and 3, the dimension of input features and linear transforms is 256 and 512 respectively. 

The hyper-parameters $\lambda_D$ and $\lambda_R$ are all set to 1.0 in our experiments. In end-to-end training phase, the initial learning rate is set as 0.1 for text spotting branches and 1.0 for information extraction branch with ADADELTA \cite{adadelta} optimization after sufficient pre-training of the former. We also decrease it to a tenth every 25 epoches for two times.


\subsection{Ablation Study}
In this section, we evaluate the influences of multiple components of the proposed framework on the EPHOIE dataset.

\begin{table*}[t]
\centering
\caption{Effect of End-to-End Optimization. \textbf{LA} indicates the whole line accuracy. \textbf{A + B} indicates the combination of optimization method A and IE structure B.}
\resizebox{170mm}{20mm}{
\begin{tabular}{cccccccc}
\toprule
\multirow{2}{*}{\textbf{Task}} & \multicolumn{2}{c}{\multirow{2}{*}{\textbf{Mesure}}} & \multicolumn{5}{c}{\textbf{Optimization Strategy}} \\ \cmidrule(l){4-8} 
                               &                   &               & Baseline       & TRIE      & E2E(Ours) + IE stage in TRIE      & E2E(Ours) + GAT      & \textbf{VIES(Ours)}      \\ \hline
\multirow{4}{*}{Detection}                                       & IoU=0.5,         & \multirow{4}{*}{F1-Score}         & 97.00                            & 97.31 & 97.36       & 97.10            & \textbf{97.48    }                      \\
                                                                 & IoU=0.6,         &                                   & 96.03                            & \textbf{96.33} & 96.25       & 96.05            & 96.15                          \\
                                                                 & IoU=0.7,         &                                   & 92.72                            & 93.02 & 92.92       & 92.48            & \textbf{93.06}                          \\
                                                                 & IoU=0.8,         &                                   & 78.89                            & 78.30 & 79.04       & 78.16            & \textbf{79.60}                          \\ \hline
\multirow{2}{*}{Recognition}                                     & \multicolumn{2}{c}{AR}                               & 96.43                            & 96.28 & 96.40       & 95.59            & \textbf{96.79}                         \\ \cline{2-8} 
                                                                 & \multicolumn{2}{c}{LA}                              & 93.98                            & 93.53 & 93.78       & 93.55            & \textbf{94.52}                        \\ \hline
\begin{tabular}[c]{@{}c@{}}Information\\ Extraction\end{tabular} & \multicolumn{2}{c}{F1-Score}                         & 80.31                            & 81.24 & 82.21       & 80.51            & \textbf{83.81}                          \\ \hline
\end{tabular}
}
\label{cm}
\end{table*}

\subsubsection{Effect of End-to-End Optimization}

To explore the effects of end-to-end optimization manner introduced by VCM and SCM, we perform the following ablation studies and the results are presented  in Table \ref{cm}.
\textbf{Baseline} denotes the gradients generated by information extraction branch are detached and cannot be back-propagated to the text spotting part. 
We select two other advanced structures -- graph attention network (GAT) \cite{gat}  similar to  \cite{GCN} and the information extraction module in TRIE \cite{TRIE}, then combine them with the optimization methods both in TRIE and our VIES for detailed comparison. 

From Table \ref{cm}, it can be seen that \textbf{VIES(Ours)} outperforms four counterparts in all of text detection, recognition and information extraction tasks by a large margin, which reveals the superiority of our framework. 
\textbf{TRIE} performs better in detection task under low IoU and information extraction task than \textbf{Baseline}, however, the performance of detection with high IoU and recognition are both evidently  reduced. This indicates that the improvement of final achievements does not always mean the overall progress of the entire  system. Compared to it, \textbf{E2E(Ours) + IE stage in TRIE} achieves comparable detection results under low IoU and significantly better performances on other counts, which fully verifies the advantage of our optimization strategy. Moreover, \textbf{VIES(Ours)} shows significant gains in all tasks over \textbf{E2E(Ours) + IE stage in TRIE} and \textbf{E2E(Ours) + GAT}, revealing both the effectiveness of modeling of our AFFM and the fact that, the co-training method needs to be built under careful considerations to take full advantage of its role.

\begin{table}[t!]
\caption{Effects of VCM and SCM. $H$ denotes the input feature sequence and $S$ denotes hidden states in the recognition branch.}
\centering
\resizebox{80mm}{8mm}{\begin{tabular}{@{}llc@{}}
\toprule
\textbf{Setting} & \multicolumn{1}{c}{\textbf{The design of VCM}} & \multicolumn{1}{c}{\textbf{F1-Score}} \\ \midrule
1)              & Adopting \textbf{RoIPooling} as $RegionPool$                 &         83.28                              \\ \hline
\textbf{Ours}    & Adopting \textbf{RoIAlign} as $RegionPool$        &       \textbf{83.81}                                \\ \bottomrule
\end{tabular}}
\resizebox{80mm}{11.5mm}{\begin{tabular}{@{}llc@{}}
\toprule
\textbf{Setting} & \multicolumn{1}{c}{\textbf{The design of SCM}} & \multicolumn{1}{c}{\textbf{F1-Score}} \\ \midrule
2)              & \begin{tabular}[l]{@{}l@{}}Taking a further decoding step  after \\ predicting token $<$END$>$  \end{tabular}              &         80.26                              \\
3)              & Extracting from $H$  with 1D convolutions       &         82.59                              \\ \hline
\textbf{Ours}    & Extracting from $S$  with 1D convolutions        &       \textbf{83.81}                                \\ \bottomrule
\end{tabular}}
\label{scm}
\end{table}

\subsubsection{Effects of VCM and SCM}
Here we conduct the following experiments to verify the effects of our VCM and SCM. We design several intuitive and effective structures for them and the results are shown in Table \ref{scm}. It totally indicates that although combining text spotting branches and IE branch is a relatively intuitive idea, how to make the best use of it needs a comprehensive design. Our VCM and SCM can maximize the benefits of end-to-end optimization.
\subsubsection{Effect of multi-source features}
\begin{table}[t!]
\caption{Effect of multi-source features.}
\centering
\begin{tabular}{ccccc}
\toprule
\multirow{2}{*}{\textbf{Setting}} & \multicolumn{3}{c}{\textbf{Source}}                                               & \multirow{2}{*}{\textbf{F1-Score}} \\ \cline{2-4}
                                  & Semantics                 & Vision                    & Location                  &                                    \\ \hline
(1)                               & \checkmark &                           &                           & 80.51                              \\
(2)                               & \checkmark & \checkmark &                           & 83.25                              \\
(3)                               & \checkmark &                           & \checkmark & 81.58                              \\
(4)                               & \checkmark & \checkmark & \checkmark & \textbf{83.81}    \\ \bottomrule
\end{tabular}
\label{ablation1}
\end{table}
We conduct the following experiments to verify the effectiveness of multi-source features in AFFM, and the results are presented  in Table \ref{ablation1}. It can be observed  that further fusion of the multi-modality representations in Setting \textbf{(4)} provides  the best performance. 

Semantic features are the most distinguishable ones for information extraction. As shown in Setting \textbf{(1)}, our method can achieve satisfactory performance by using only semantic features. Moreover, these features are provided from our recognition branch, which may be more effective than the re-extracted ones under traditional routines.

Visual features such as fonts and colors which containing rich semantic clues is also crucial. This brings significant performance gains, which can be observed in Setting \textbf{(2)}. When semantics of different texts are highly similar, visual features play the decisive role. 

Note that, introducing spatial features in Setting \textbf{(3)} outperforms Setting \textbf{(1)} slightly, revealing that the shape and location of texts also plays a critical role in representing semantic meanings. Through the adaptive feature fusion process, the expressive features mentioned above belong to different individuals are allowed freely attending to all the others, which enables modeling both inter- and intra- segment relations in a homogeneous manner. The negative effects of errors from different sources can also be mitigated here.

\subsection{Comparison with the State-of-the-Arts}
To comprehensively evaluate our framework, we compared it with several state-of-the-art methods. It is notable that, we re-implement them based on the original papers or source codes if available on open-source platforms.

\begin{table*}[]
\centering
\caption{Performance (F1-Score) comparison of the state-of-the-art  algorithms on  the EPHOIE dataset. \textbf{Ground Truth} means using ground truth bounding boxes and texts as inputs for information extraction branch, and \textbf{End-to-End} denotes using same predictions from text spotting branches instead.}
\resizebox{175mm}{23mm}{
\begin{tabular}{@{}clccccccccccc@{}}
\toprule
\multirow{2}{*}{\textbf{Setting}}      & \multicolumn{1}{c}{\multirow{2}{*}{\textbf{Method}}} & \multicolumn{11}{c}{\textbf{Entities}}                                                                                                                                                                                                                                                                                    \\ \cmidrule(l){3-13} 
                                       & \multicolumn{1}{c}{}                                 & Subject        & Test Time       & Name           & School          & \begin{tabular}[c]{@{}c@{}}Examination\\ Number\end{tabular} & \begin{tabular}[c]{@{}c@{}}Seat\\ Number\end{tabular} & Class          & \begin{tabular}[c]{@{}c@{}}Student\\ Number\end{tabular} & Grade          & Score          & \textbf{Mean}  \\ \midrule
\multirow{5}{*}{\textbf{\begin{tabular}[c]{@{}c@{}}Ground\\ Truth\end{tabular}}} &  \cite{blstmcrf}                                           & 98.51          & \textbf{100.0} & 98.87          & 98.80           & 75.86                                                        & 72.73                                                 & 94.04          & 84.44                                                    & 98.18          & 69.57          & 89.10          \\
                                       &  \cite{GCN}                                                  & 98.18          & \textbf{100.0} & 99.52          & \textbf{100.0} & 88.17                                                        & 86.00                                                 & 97.39          & 80.00                                                    & 94.44          & 81.82          & 92.55          \\
                                       & GraphIE \cite{Graphie}                                              & 94.00          & \textbf{100.0} & 95.84          & 97.06           & 82.19                                                        & 84.44                                                 & 93.07          & 85.33                                                    & 94.44          & 76.19          & 90.26          \\
                                       & TRIE \cite{TRIE}                                                 & 98.79          & \textbf{100.0} & 99.46          & 99.64           & 88.64                                                        & 85.92                                                 & 97.94          & 84.32                                                    & 97.02          & 80.39          & 93.21          \\ \cmidrule(l){2-13}
                                       & \textbf{VIES(Ours)}                                 & \textbf{99.39} & \textbf{100.0} & \textbf{99.67} & 99.28           & \textbf{91.81}                                               & \textbf{88.73}                                        & \textbf{99.29} & \textbf{89.47}                                           & \textbf{98.35} & \textbf{86.27} & \textbf{95.23} \\ \hline
\multirow{4}{*}{\textbf{End-to-End}}    &  \cite{blstmcrf}                                         & 82.08          & 89.95           & 72.61          & 83.29           & 62.18                                                                 & 64.56                                                          & 66.87          & 63.68                                                             & 81.17          & 53.09          & 71.95          \\
                                                                                 &  \cite{GCN}                                                  & 84.12          & 90.61           & 78.35 & 87.25           & 68.60                                                                 & 64.45                                                          & 71.56          & 68.39                                                             & 82.19          & 55.22          & 75.07     \\
                                                                                 & TRIE \cite{TRIE}                                                 & 85.92          & 92.20           & 85.94          & 91.92           & 73.63                                                                 & 69.01                                                          & 79.91          & 78.00                                                             & \textbf{83.82} & 62.74          & 80.31           \\\cmidrule(l){2-13}
                                       & \textbf{VIES(Ours)}                                 & \textbf{86.14} & \textbf{93.50}  & \textbf{90.35} & \textbf{95.47}  & \textbf{77.72}                                               & \textbf{76.05}                                        & \textbf{85.65} & \textbf{81.05}                                           & 83.49          & \textbf{68.62} & \textbf{83.81} \\ \bottomrule
\end{tabular}
}
\label{ephoie}
\end{table*}

\subsubsection{Results on EPHOIE Dataset}
\begin{figure}[t!]
\centering
\subfigure[]{
\begin{minipage}[]{0.95\linewidth}
\centering
\includegraphics[width=8cm,height=1.8cm]{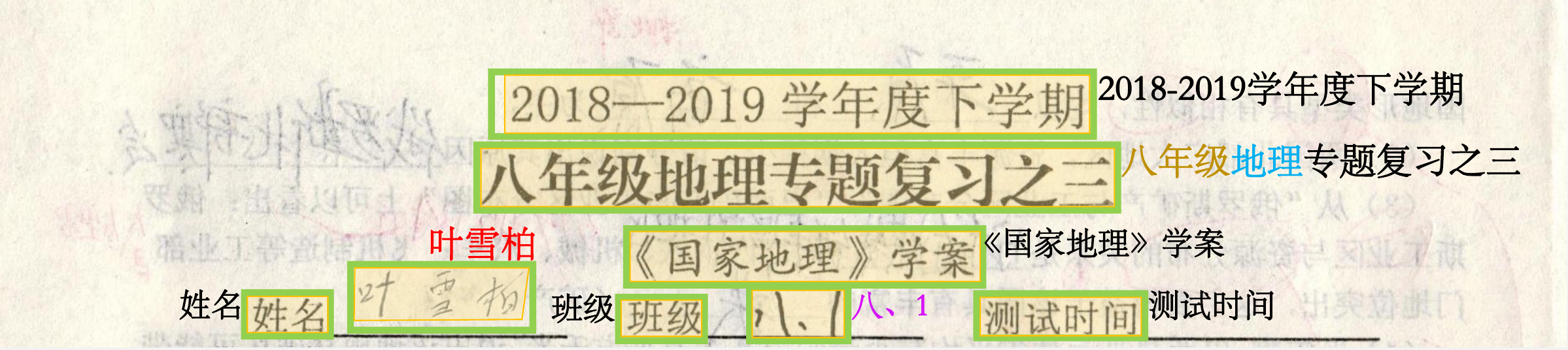}\\
\end{minipage}
}%
\quad
\subfigure[]{
\begin{minipage}[]{0.95\linewidth}
\centering
\includegraphics[width=8cm,height=1.8cm]{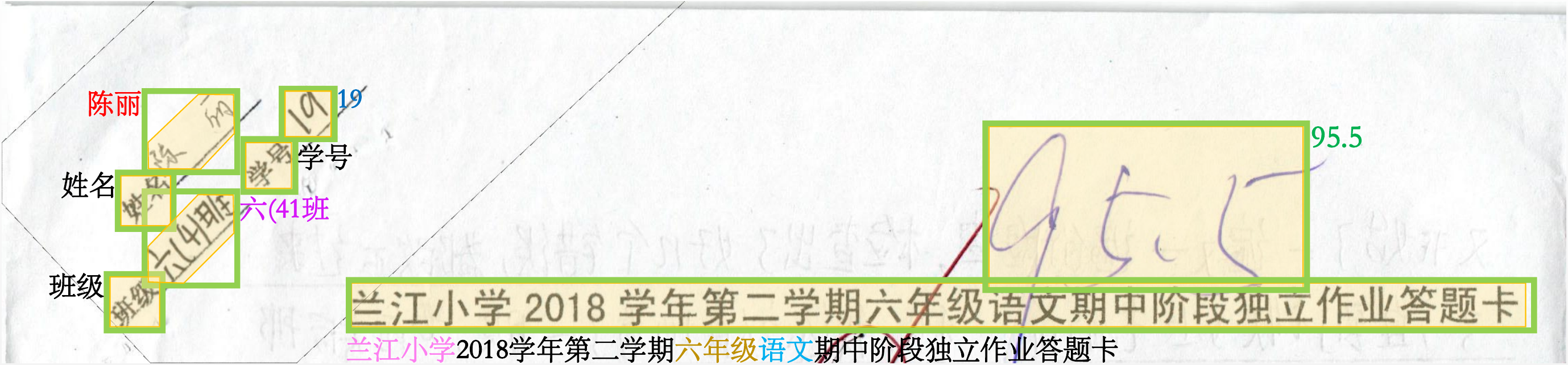}\\
\end{minipage}
}%
\caption{Examples of prediction results of VIES on EPHOIE. Different colors denotes different entities.}\label{fig:res}
\end{figure}

As shown in Table \ref{ephoie}, our method exhibits superior performance on EPHOIE.
 \cite{GCN}, TRIE \cite{TRIE} and VIES which introduce multimodal representations outperform counterparts by significant margins. Under the \textbf{End-to-End} setting where the OCR results are less accurate, the robustness of our pipeline is more evident. Its reasonable design effectively reduces the negative effects caused by text spotting errors. Some examples of qualitative results of VIES are shown in Figure \ref{fig:res}.

\begin{table}[t!]
\caption{Performance comparison of the state-of-the-art algorithms on SROIE dataset. $\dagger$ indicates the result is reported in  \cite{TRIE}. \textbf{Competition} shows the performance  of  the  top  three  methods  during ICDAR 2019  SROIE  Competition which inevitably introduced techniques such as model ensemble and complex post-processing.}
\centering
\resizebox{75mm}{32mm}{\begin{tabular}{@{}llc@{}}
\toprule
\textbf{Setting}                     & \textbf{Method}             & \textbf{F1-Score} \\ \midrule
\multirow{7}{*}{\textbf{\begin{tabular}[l]{@{}l@{}}Ground\\ Truth\end{tabular}}}  &  \cite{blstmcrf}$\dagger$ & 90.85    \\
                             & LayoutLM \cite{Layoutlm}            & 95.24    \\
                             &  \cite{GCN}                &     95.10      \\
                             & PICK \cite{PICK}                & 96.12    \\
                             & TRIE \cite{TRIE}                & \textbf{96.18}    \\ \cmidrule(l){2-3}
                             & \textbf{VIES(Ours)}         &    96.12      \\\hline
\multirow{6}{*}{\textbf{End-to-End}}     
                             & NER \cite{ner}$\dagger$                & 69.09    \\
                             & Chargrid \cite{Chargrid}$\dagger$            & 78.24    \\  &  \cite{blstmcrf}          &    78.60      \\
                             & \cite{GCN}                &       80.76   \\
                             & TRIE \cite{TRIE}                & 82.06    \\\cmidrule(l){2-3}
                             & \textbf{VIES(Ours)}         & \textbf{91.07}    \\\hline
\multirow{3}{*}{\begin{tabular}[l]{@{}l@{}}\textbf{Competition}\\  \cite{sroie}\end{tabular}} & Rank 1              & 90.49    \\
                             & Rank 2              & 89.70     \\
                             & Rank 3              & 89.63    \\ \bottomrule
\end{tabular}
}
\label{sroie}
\end{table}

\subsubsection{Results on SROIE Dataset}
The results of experiments on SROIE dataset are shown in Table \ref{sroie}. Our method achieves competitive results under \textbf{Ground Truth} setting and outperforms the state-of-the-art results by significant margins (\textbf{from 82.06 to 91.07}) under \textbf{End-to-End} setting. The methods in \textbf{Competition} may inevitably introduce model ensemble techniques for each tasks and complex post-processing. However, our VIES achieves even better  results using only a single framework with light-weight network structures. And we only introduce simple regularizations to correct the format of \textit{Total} and \textit{Date} results.

Compared with EPHOIE, the layout of scanned receipts is relatively fixed, the font style is less changeable and there exists  less noise in the background. In such a relatively simple scenario, the superiority of our method is further confirmed.

\section{Conclusion}
In this paper, we propose a robust visual information extraction system (VIES) towards  real-world  scenarios,  which  is  an  unified  end-to-end trainable framework for simultaneous text detection, recognition and information extraction.  

Additionally, we propose a fully-annotated dataset called EPHOIE, which is the first Chinese benchmark for both OCR and VIE tasks. Extensive experiments demonstrate that  our VIES achieves superior performance on the EPHOIE  dataset and  has  9.01\%    F-score  gains  campared with the previous state-of-the-art methods on 
the widely used SROIE dataset under the end-to-end information extraction scenario. 

Visual information extraction is a challenging task in the cross domain of natural language processing and computer vision. 
Many issues have not been well addressed, including complex layouts and background, over-reliance on complete annotations and continuous accumulation of errors. Therefore, it remains an open research problem and deserves more attention and further investigation.

\section{ Acknowledgments}
This research is supported in part by NSFC (Grant No.: 61936003, 61771199),  GD-NSF (No. 2017A030312006), the National Key Research and Development Program  of China (No. 2016YFB1001405), Guangdong Intellectual Property Office Project (2018-10-1), and Guangzhou Science, Technology and Innovation Project (201704020134).


\end{document}